\documentclass[conference, letterpaper]{IEEEtran}

\usepackage{amssymb}
\usepackage{algorithm}
\usepackage{algorithmic}
\usepackage{setspace}
\usepackage{flushend}

\usepackage{subcaption}

\ifCLASSINFOpdf
   \usepackage[pdftex]{graphicx}
\else
\fi

\usepackage[cmex10]{amsmath}
\usepackage{color}
\usepackage{fancyhdr}

\renewcommand{\thispagestyle}[2]{} 

\fancypagestyle{plain}{
        \fancyhead{}
        \fancyhead[C]{first page center header}
        \fancyfoot{}
        \fancyfoot[C]{first page center footer}
}
\begin{document}

\title{MEBoost: Mixing Estimators with Boosting for Imbalanced Data Classification }

\author{
\IEEEauthorblockN{Farshid Rayhan, Sajid Ahmed, Asif Mahbub, Md. Rafsan Jani, \\Swakkhar Shatabda, Dewan Md. Farid and Chowdhury Mofizur Rahman}
\IEEEauthorblockA{Department of Computer Science \& Engineering, United International University, Bangladesh\\  
Email: frayhan133057@bscse.uiu.ac.bd}}

\maketitle

\begin{abstract}
Class imbalance problem has been a challenging research problem in the fields of machine learning and data mining as most real life datasets are imbalanced. Several existing machine learning algorithms try to maximize the accuracy classification by correctly identifying majority class samples while ignoring the minority class. However, the concept of the minority class instances usually represents a higher interest than the majority class. Recently, several cost sensitive methods, ensemble models and sampling techniques have been used in literature in order to classify imbalance datasets. In this paper, we propose MEBoost, a new boosting algorithm for imbalanced datasets. MEBoost mixes two different weak learners with boosting to improve the performance on imbalanced datasets. MEBoost is an alternative to the existing techniques such as SMOTEBoost, RUSBoost, Adaboost, etc. The performance of MEBoost has been evaluated on 12 benchmark imbalanced datasets with state of the art ensemble methods like  SMOTEBoost, RUSBoost, Easy Ensemble, EUSBoost, DataBoost. Experimental results show significant improvement over the other methods and it can be concluded that MEBoost is an effective and promising algorithm to deal with imbalance datasets. The python version of the code is available here: https://github.com/farshidrayhanuiu/                   
   
\end{abstract}


\IEEEpeerreviewmaketitle

\section{Introduction}

In supervised learning, Machine learning is a process of identifying new or unknown samples employing classification algorithms based on a group of instances \cite{Farid_ESWA_vol64,Farid_ESWA_vol41,Farid_ESWA_vol40,Farid_Benelearn}. In real world, datasets are often high dimensional, multi class and imbalanced. Typical machine learning algorithms often fail to get a good classification accuracy on these datasets. There are two types of methods to deal with imbalanced datasets : (a) internal  and  (b) external method. Internal method modifies the preexisting algorithm in order to reduce their sensitiveness to the imbalance ratio of the dataset. External methods apply various data balancing techniques to reduce the imbalance ratio of the dataset.
 
Primarily there are two types of sampling methods to modify the original distribution of the dataset, they are over sampling and under sampling. Under sampling method reduces instances from the major class based on some intuition or just randomly. Several mainstream under sampling methods are neighborhood cleaning rule \cite{laurikkala2001improving}, 
near miss \cite{mani2003knn}, clustered under sampling \cite{yen2009cluster,rayhan2017idti}, One sided selection \cite{kubat1997addressing}. Over sampling methods perform in the opposite manner of that of under sampling. In stead of removing instances from the major class it generates samples of minority class using the minority class itself using various techniques such as AdaSyn \cite{He}, SMOTE \cite{Chawla} or randomly. Both of these techniques have some draw backs. Under sampling method has the potential to lose informative data as it reduces samples from major class data. On the other hand Over sampling generates samples from the minority which creates the potential risk of over fitting \cite{Sun}.

In the case of imbalanced datasets, the minority class instances are often outnumbered. Even though  the concept they represent is usually more important than that of the major class.  Traditional machine learning algorithms for data mining like k-nearest neighbors \cite{Farid_ESWA_vol64}, decision tree \cite{Farid_ESWA_vol40}, Support Vector Machine  \cite{cortes1995support}, Random Forest \cite{liaw2002classification} usually tries to maximize the accuracy of classification rate while ignoring miss classification cost of minority class. Various  cost sensitive methods has been proposed to deal the class imbalance problem. Cost sensitive learning applies different costs for misclassification errors to each classes. The goal is to set costs in such way that misclassification cost for minority class will be high and low for majority class. Stable classification is hard to find using cost sensitive method as it is very difficult to set the correct misclassification cost for each class. Ensemble models like bagging and boosting are typically used for imbalanced classification \cite{He,Yanmin-Sun}. Ensemble classifier is a type of algorithm where multiple learners are used to improve the performance of individual classification by combining hypotheses of each learner \cite{Farid_ESWA_vol40}.  
 
 In this paper, we present a new boosting algorithm called MEBoost, mixing two weak estimators alternately on the training set. As weak estimators, we have used decision tree and extra tree classifier. By this way are we taking advantage of both learners while avoiding the limitations by using a single base classifier in a boosting model. We tested the performance of MEBoost with other state-of-the-art boosting classifiers like Adaboost, RUSBoost, SMOTEBoost, DataBoost, EUSBoost, Easy ensemble on 12 standard benchmark imbalanced datasets. From the experimental results, it can be validated that using two learners such as decision tree and extra tree classifier alternately with Adaboost significantly improves over the performance of other algorithms and is a promising technique to handle class imbalance problem.   

Rest of the paper is organized as follows, Section \ref{related_work} represents related work; Section \ref{MEBoost} presents the details of the proposed MEBoost algorithm; Section \ref{experiments} shows the experimental results and Section~\ref{conclusion} concludes the paper.                                       

\section{Related work \label{related_work}}

\begin{figure}[!htb]
\centering
\vspace{5mm}
\includegraphics [scale=00.3]{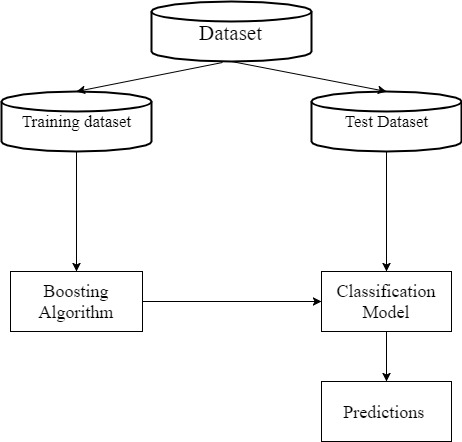}
\vspace{1mm}
\caption{Boosting for classifying imbalanced data.}
\label{sampling_Boosting}
\end{figure}

Through out last decade various sampling methods, ensemble methods, ensemble methods based on bagging and boosting has been the prime focus for dealing the classification problem with class imbalance datasets. Fig.~\ref{sampling_Boosting} depicts a general sketch of applying boosting algorithm to classification problem. 

Several ensemble methods are proposed in the literature to handle imbalanced datasets \cite{He,Yanmin-Sun}. An ensemble method was proposed by Sun et al. \cite{Sun} which converts a imbalance binary class problem to multiple  learning process. The proposed method divided the majority class instances into several sub datasets. Here each sub datasets holds almost same number of minority class instances. Thus several balances datasets were created and used to create a binary classifiers. Then a combination of those classifiers were used to learn an ensemble classifier.     

Boosting is a meta classifier that combines the predictions of multiple base estimators and uses a voting technique for classification. It assigns weights to instances based on how arduous they are to classify. Thus it sets high weights to to hard instances. Here each estimator's contributed weight is used by the next estimator. Then based on the base learner's predictive accuracy weights are assigned to it. There weights are taken in consideration for new instance prediction. Though boosting was not intended for class imbalance problem, due to this characteristic it has become quite ideal for class imbalance problem. 

RUSBoost \cite{Seiffert} is a hybrid boosting algorithm using Adaboost with random under sampling as sampling method. From an imbalanced data random under sampling randomly removes instances from major class in each iteration. A Adaboost \cite{freund1996experiments} was used with random under sampling to create the RUSBoost algorithm. Similarly SMOTEBoost was created using Adaboost and a over sampling technique called SMOTE. This method was proposed in \cite{Chawla_SMOTEBoost}. It over-samples the minority class instances using an over sampling technique called SMOTE \cite{Chawla}. By employing $k$ nearest neighbors of minority class, synthetic instances are generated by operating in feature space. The above mentioned methods which uses sampling inside Adaboost showed impressive performance in terms of area under Receiver Operating Characteristic (ROC) curve.  Investigation on the behavior of SMOTEBoost was performed by Blagus and Lusa \cite{Blagus} on imbalanced datasets with high dimensions. Here dataset with high dimension means where there are more features than the instances. They came to the conclusion that, as SMOTE biases the classifier towards minority class, it is necessary to do feature selection. 

De Souza et al. proposed a new dynamic \cite{de2011extending} Adaboost algorithm where 10 different estimators were used alternately in each iteration. As Adaboost keeps the better estimators and discards estimator with high error, by allowing 10 different estimators alternately it reduces the burden of the user to choose a learner. This algorithm does not follow the weak learner concept as it uses estimators like Random forest, SVM, Neural Network thus also making it highly computationally expensive.   An evolutionary ensemble boosting algorithm was proposed by Galar et al. \cite{Galar}. It uses evolutionary under-sampling method thus called EUSBoost. This algorithm creates several sub datasets are generated by random under sampling method in order to find the best under-sampled from the original dataset. EUSBoost was also built based on AdaBoost algorithm \cite{freund1996experiments}. 

DataBoost algorithm or DataBoost-IM method was presented by	Hongyu Guo 
\cite{guo2004learning}. He proposed an ensemble model which uses data generation. In this algorithm, hard majority and minority class instances are identified during the execution of boosting. Then those hard examples are chosen separately and used to create synthetic instances of respective class. After that those created instances are added to the main dataset. Easy Ensemble is an ensemble method was proposed by Xu-Yung Liu \cite{liu2009exploratory}. They create several subsets of majority class instances. Then using each of those datasets it trains a learner. These subsets are created using random under sampling. However, it creates several sub datasets to overcome the main limitation of random under sampling which is it discard instances from majority class randomly regardless of its importance.

\section{MEBoost Algorithm \label{MEBoost}}

Most of the boosting algorithms discussed in Section~\ref{related_work} used a single weak estimator to create the ensemble model. In our proposed method MEBoost, instead of using a single estimator we use two different estimators alternately. For each iteration it either uses decision tree or extra tree classifier as its learner. By doing this the algorithm is taking the benefits of both classifiers. The algorithm is also discarding learners with poor performance by design of the boosting procedure. MEBoost algorithm does not perform any sampling on the train set. For each iteration we use decision tree and Extra tree classifier alternately on the train dataset. Decision trees are built from train set using information entropy \cite{quinlan1996improved}. The train data $S = { s_1, s_2, s_3, s_4,... }$ is classified samples. Sample $S_i$ consists of vector $X = { x_1,_i, x_2,_i, x_3,_i, x_4,_i }$. Here $X_ij$ is a representation of samples' feature or values of the attribute. For each node an feature/attribute is choose in such way that it splits the train dataset into subsets of each class most effectively. Extra tree is a randomized tree classification algorithm. While looking for best split in order to separate instances of a node into groups, Extra tree draws random splits for each number of feature randomly selected and among them the best split is chosen \cite{geurts2006extremely}. Extra Tree Classifier acts like a decision tree if the number of randomly selected features are 1.

The pseudo-code of our proposed method MEBoost is given in Algorithm~\ref{Algo1}. It is a modification of the basic Adaboost classifier. The number of base classifiers is not restricted here. However, at each iteration of the algorithm, MEBoost tests each of the weak estimators learned and discards if it fails to be a weak classifier or its error rate is greater or equal to 0.5. The meta classifier is tested on the test data kept apart and stores the best combination according to the auROC score. It keeps adding weak learners to the model until there are no significant change in the auROC on the test data. 

\begin{figure}[h]
	\centering
	\vspace{5mm}
	\includegraphics [scale=0.45]{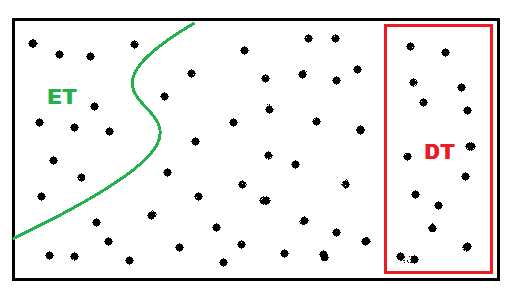}
	
	\vspace{2mm}
	\caption{Instances covered by different tree algorithms}
	\label{diagram}
\end{figure}

The intuition behind this idea is that tree algorithms like extra tree and decision tree usually are better suited for boosting scheme cause of their instability. In a particular dataset, any number of SVM is more likely to create similar decision boundaries. But there is a good possibility that the on that dataset each tree algorithms will generate different trees in different ways and cover different sub-spaces of the total dataset. So as they cover different sub-spaces, combing them under the boosting scheme is a excellent recipe for good classification algorithm. In order to further maximize the diversity, we used 2 different tree algorithms, ie Extra tree and decision tree, under the boosting scheme. In Fig \ref{diagram}, the rectangular represents the whole dataset and each black dot represents an instance. Here the dots inside red box represents the part of dataset that has be explored by a Decision Tree(DT). Similarly the green boundary represents the part of dataset that has be explored by a Extra Tree(ET). By using them inside a boosting mechanism, we take the instability characteristic of the tree algorithms as an advantage for better classification results.

Convergence of the algorithm depends on a stagnation window parameter $W$. In this paper, we kept $W=10$. Therefore after the best combination found it will continue up to adding $W$ estimators in the model and if there are no significant improvement in the test score then the ensemble model will return the best meta classifier learned. This early stopping criteria\cite{yao2007early} was first used by B{\"u}hlmann in 2003 \cite{buhlmann2003boosting} and were later studied by Jiang \cite{jiang2004process}. MEBoost algorithm is a combination of De Souza's alternative estimator usage and B{\"u}hlmann's early stopping criteria in an Adaboost algorithm.     



\begin{algorithm}
	\caption{MEBoost} \label{Algo1}
	\textbf{Input:} Imbalanced data, $D$, Window size $W$ \\
	\textbf{Output:} An ensemble model $H$. \\  
	\textbf{Method:}
	\begin{algorithmic}[1]
		\STATE set $t$ to $0$, $score_{best}$ to $0$ and $nonImproving$ to $0$ 
		\STATE initialize weight, $w_{i}^{t}$ to $\frac {1} {|D|}$ for each $x_i \in D$;
		\WHILE{$true$}
		\STATE increase $t$;
		\STATE select the estimator tree type $T$
		\STATE learn estimator $h^{(t)}$ of type $T$
		\STATE compute the error rate of $h^{(t)}$, $error(h^{(t)})$
		\IF{$error(h^{(t)})$ $\geq 0.5$ }
		\STATE go back to step 4 and try again;
		
		\ENDIF
		\FOR{each $x_{i} \in D$ }
		\STATE update weights $w_i^t$
		\ENDFOR
		\STATE $\alpha^{t}=\frac{1}{2}log_e\frac{1-error(h^{(t)})}{error(h^{(t)})}$
		\STATE learn meta classifier $H^{t}=sign(\sum \alpha^t h^{(i)})$
		\STATE $score$ = $auROC score(H^{t},X_{test},Y_{test} )$
		\IF  { $score_{top}$ $\leq$ $score$  }
		\STATE $score_{best}$ $=$ $score$
		\STATE $H_{best}=H^{t}$ 
		\ENDIF
		\IF {$score$ $\leq$ $score_{top}$ }
		\STATE $nonImproving=nonImproving+1$
		\IF { $nonImproving=W$  }
		\STATE $H$ = $H_{best}$
		\STATE break
		\ENDIF
		\ENDIF   
		\ENDWHILE
	\end{algorithmic}
\end{algorithm}

\section{Experimental Results}
\label{experiments}
This section presets the experimental results and analysis of the performance of the MEBoost algorithm. 

\subsection{Benchmark Datasets}

Datasets with different imbalance ratio were chosen from KEEL-dataset repository \cite{KEEL-dataset}. The following TABLE \ref{datasets} presents a summary of the datasets. Each of the datasets imbalanced with imbalance ratio lies in the range from 1.87 up to 41.03.   

\begin{table}[h!]\footnotesize
	  \centering
	  \caption{Datasets description.}\label{datasets}
\begin{tabular}{ | l | l | l | l | l | }
	\hline
	Name   				& Imbalance ratio & instances & features  \\ \hline
	pima 				& 1.87 & 768 & 8 \\ \hline
	glass-0-1-2-3vs4-5-6& 3.2  & 214 & 9 \\ \hline
	newthyroid2  		& 5.14 & 215 & 5 \\ \hline
	newthyroid1  		& 5.14 & 215 & 5 \\ \hline
	segment0  			& 6.02 & 2308& 19 \\ \hline
	glass6  			& 6.38 & 214 & 9 \\ \hline
	yeast-2\_vs\_4  	& 9.08 & 514 & 8 \\ \hline
	page\_blocks-1-3-vs 4& 15.86 & 472 & 10 \\ \hline
	glass5  			& 22.78& 214 & 9 \\ \hline
	$yeast4$  			& 28.1 & 1484& 8 \\ \hline
	$yeast5$  			& 32.72& 1484& 8 \\ \hline
	$yeast6$  			& 41.03& 1484& 8 \\ \hline
\end{tabular}
\end{table}

\begin{table*}[!htb] 
  \centering\label{Mean_Result1}
  \caption{Average performance of the AdaBoost, EUSBoost, EasyEnsemble, SMOTEBoost, RUSBoost, DataBoost and MEBoost methods on 12 imbalanced datasets.}\label{Mean_Result1}
	\begin{tabular}{ | l | l | l | l | l | l |l |l |}
		\hline\label{Mean_Result1}
		Dataset  			& Adaboost & EUSBoost     & EasyEnsemble& SMOTEBoost& RUSBoost 		& DataBoost & MEBoost  \\  \hline
		pima 			& 0.68 	   & \textbf{0.73}& 0.71 		& 0.73 		& \textbf{0.82} & 0.72 		& 0.71		\\ \hline
		glass-0-1-2-3 	& 0.89 	   & 0.91 		  & 0.9   		& 0.91 		& 0.96 			& 0.9		& \textbf{0.98} \\ \hline
		newthyroid2 	& 0.95 	   & 0.94 		  & 0.95  		& 0.9 		& 0.98 			& 0.93		& \textbf{0.99} \\ \hline
		newthyroid1 	& 0.94     & 0.96 		  & 0.96  		& 0.98 		& 0.98 			& 0.97		& \textbf{0.99} \\ \hline
		segment0 		& 0.96 	   & 0.97 		  & 0.98  		&\textbf{0.99}& \textbf{0.99}	& \textbf{0.99}	& \textbf{0.99} \\ \hline
		glass6 			& 0.84 	   & 0.9          & 0.88  		& 0.85 		& 0.89 			& 0.88 		& \textbf{0.99} \\ \hline
		yeast-2\_vs\_4 	& 0.88     & 0.9          & 0.89  		& 0.89 		& 0.94 			& 0.89		& \textbf{0.98} \\ \hline
		page\_blocks 	& 0.92 	   & 0.98         & 0.93  		& 0.95 		& 0.97 			& 0.9		& \textbf{0.99} \\ \hline
		glass5 			& 0.97 	   & 0.97         & 0.98  		& 0.98 		& 0.97 			& 0.98		& \textbf{0.99} \\ \hline
		$yeast4$ 		& 0.62 	   & 0.85         & 0.78  		& 0.71 		& 0.89 			& 0.75		& \textbf{0.91} \\ \hline
		$yeast5$ 		& 0.81 	   & 0.94 		  & 0.95  		& 0.94 		& 0.96 			& 0.85		& \textbf{0.99} \\ \hline
		$yeast6$ 		& 0.74 	   & 0.82         & 0.84  		& 0.81 		& 0.93 			& 0.87		& \textbf{0.95} \\ \hline
	\end{tabular}

\end{table*}

\subsection{Evaluation Metrics}

Several evaluation metrics are used in the literature to measure the performance of classification algorithms. In this paper we used, area under Receiver Operating characteristic curve (auROC) as the comparison metric that has been widely used as the standard for comparison of performance in the literature of imbalanced datasets. ROC curve is a representation of best decision boundaries for cost between  true positive rate (TPR) and false positive rate (FPR). ROC curve plots TPR against FPR. TPR and FPR are defined as following:    

\begin{equation}
TPR =  \frac{TP}{TP+FN} 
\end{equation}

\begin{equation}
FPR= \frac{FP}{FP+TN}
\end{equation}

Here, TP denotes the number of positive samples correctly classified, TN denotes the number of negative samples correctly classified, FP denotes the number of negative samples incorrectly classified  and FN denotes the number of positive samples correctly classified by the estimator. A point on auROC curve is limited between $0.0$ up to $1$, where $0.0$ means all instances are misclassified and $1$ means all positive instances are classified correctly. The line $y = x$ is the minimum threshold as that line represents the scenario where classes are  randomly guessed. Area Under the ROC Curve is very useful as performance metric for class imbalance problems. Because it doesn't depend on decision criterion selected and prior probabilities. A dominant relationship can be established between classifiers using AUC comparison.

\subsection{Results}
\subsubsection{MEBoost vs other boosting algorithms }
\noindent
We compared the performance of MEBoost with that of RUSBoost, AdaBoost, Easy Ensemble,  DataBoost, SMOTEBoost and EUSBoost methods. Classification performance on the datasets were measured in terms of auROC. For the other methods C4.5 decision tree was used as  base learner. In case of MEBoost, our proposed method, we used C4.5 and Extra Tree classifiers. Keel-dataset \cite{KEEL-dataset} repository's implementation was used for the DataBoost, AdaBoost, RUSBoost, SMOTEBoost, EUSBoost and Easy Ensemble algorithms. All the datasets were spited in 3 sections : Train set, Test set and Validation set. Validation set contains $5\%$ of the dataset. The classifier was trained and tested on train and test sets using 5 fold cross validation. Mean auROC scores on validation set from 10 experiments are shown in TABLE \ref{Mean_Result1}.

TABLE \ref{Mean_Result1} shows the performance of the MEBoost classifier against other state of the art Boosting algorithms such as Adaboost, EUSBoost, EasyEnsemble, SMOTEBoost, RUSBoost and DataBoost. MEBoost was able to achieve highest auROC score in all the datasets except pima where RUSBoost achieved the highest  value. Note that the imbalance ratio of pima dataset is the lowest among all datasets.

\subsubsection{Multiple estimator vs single estimator  }
\noindent

We also compared the performance of MEBoost with boosting using single base estimators like Decision tree, Extra tree, Random forest, Support Vector Machine. Table~\ref{Result2} presents the results for this experiment. Here we used an AdaBoost algorithm. As a learner we used decision tree, extra tree, Support vector machine, Random Forest and they have been compared with our proposed method where we use multiple estimators, decision tree and extra tree alternately. The results in Table~\ref{estimator_Result} shows that the benefit of the usage of multiple estimators is promising. Even though in some cases the proposed method were unable to obtain the best result, training a learner like decision tree or extra tree takes much less computational power than training a support vector machine(SVM) or random forest. In dataset glass-0-1-2-3 and yeast6 MEBoost and Random Forest estimator achieved the highest auROC score. Similarly in dataset newthyroid2 and yeast5 MEBoost and Random Forest estimator achieved the top auROC score. In newthyroid1 and segment0 MEBoost, Random Forest estimator and SVM estimator achieved the highest auROC score together. Computation wise Support Vector Machine and Random are far more complex than Extra tree and decision tree together. So even though they provided similar score MEBoost is more preferable considering computational cost.

\begin{table}[!htb]\footnotesize\label{Result2}
	\centering
	\caption{Performance comparison of MEBoost with Adaboost using different single base estimators on 12 imbalanced datasets.}\label{estimator_Result}
	\begin{tabular}{ | l | c | c | c | c | c |}
		\hline\label{Result2}
		&Decision&Extra&Random&&\\
		Dataset&tree & tree 	& forest & SVM 			& MEBoost\\ \hline
		pima 			& 0.68 			& 0.64			& 0.66 			&0.69			& 0.71 \\ \hline
		glass-0-1-2-3 	& 0.89			& 0.97 			& \textbf{0.98}	&0.96			& \textbf{0.98} \\ \hline
		newthyroid2 	& 0.95 			& 0.97 			& 0.98 			&\textbf{0.99}	& \textbf{0.99} \\ \hline
		newthyroid1 	& 0.94			& 0.98 			& \textbf{0.99}	&\textbf{0.99}	& \textbf{0.99} \\ \hline
		segment0 		& 0.96			& \textbf{0.99}	& \textbf{0.99}	&\textbf{0.99}	& \textbf{0.99} \\ \hline
		glass6 			& 0.84			& 0.97 			& 0.93 			&0.93			& \textbf{0.99} \\ \hline
		yeast-2\_vs\_4 	& 0.88 			& 0.95 			& 0.95			&0.91			& \textbf{0.98} \\ \hline
		page\_blocks 	& 0.92 			& 0.96 			& 0.98			&0.98			& \textbf{0.99} \\ \hline
		glass5 			& 0.97 			& 0.97			& 0.95 			&0.93			& \textbf{0.99} \\ \hline
		$yeast4$ 		& 0.62 			& 0.92 			& 0.8 			&0.89			& \textbf{0.91} \\ \hline
		$yeast5$ 		& 0.81 			& 0.97 			& 0.98 			&\textbf{0.99}	& \textbf{0.99} \\ \hline
		$yeast6$ 		& 0.74 			& 0.91 			& \textbf{0.95}	&0.94			& \textbf{0.95} \\ \hline
	\end{tabular}
\end{table}

\begin{figure*}
\begin{tabular}{ccc}
\includegraphics [width=0.33\textwidth]{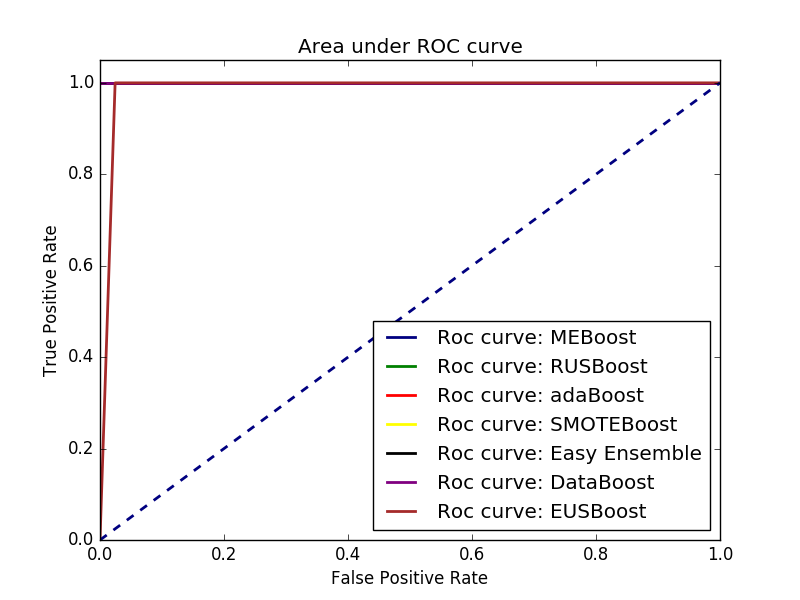}& \includegraphics [width=0.33\textwidth]{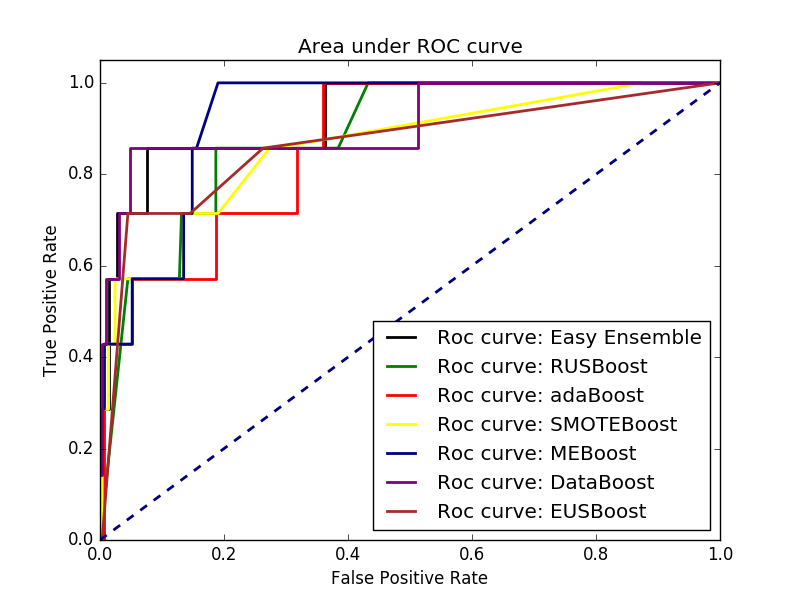}&	\includegraphics [width=0.33\textwidth]{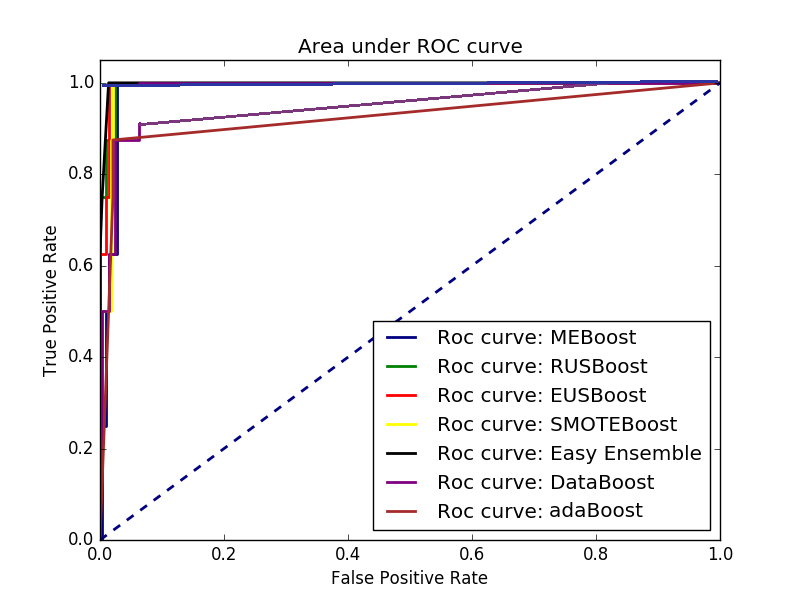}\\
(a) & (b) & (c)\\

\includegraphics [width=0.33\textwidth]{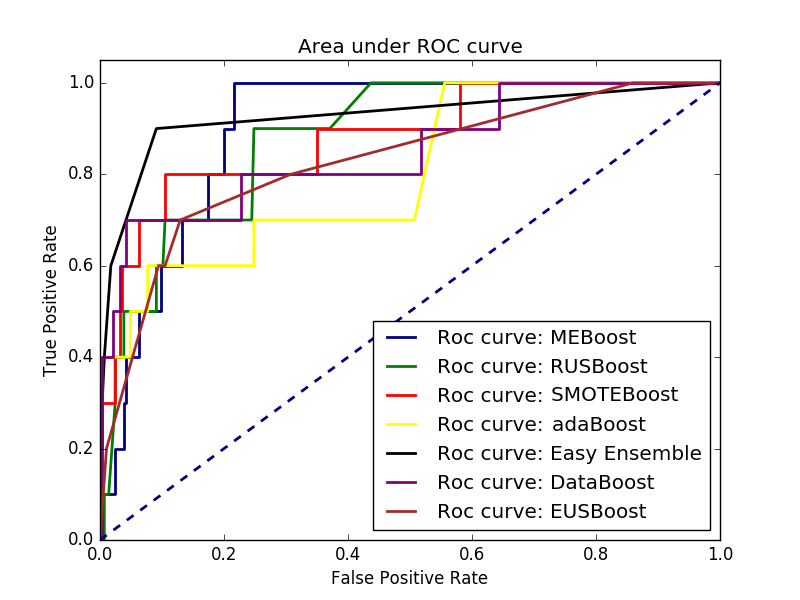}& \includegraphics [width=0.33\textwidth]{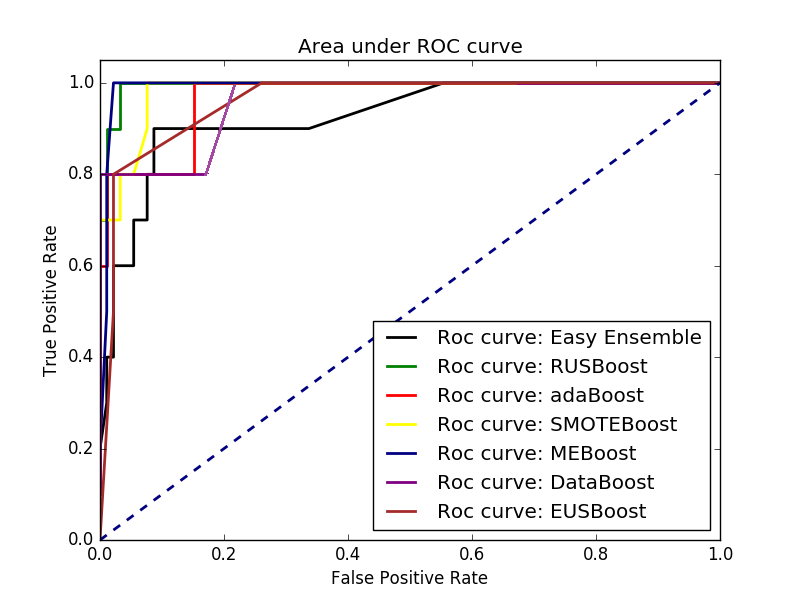}&	\includegraphics [width=0.33\textwidth]{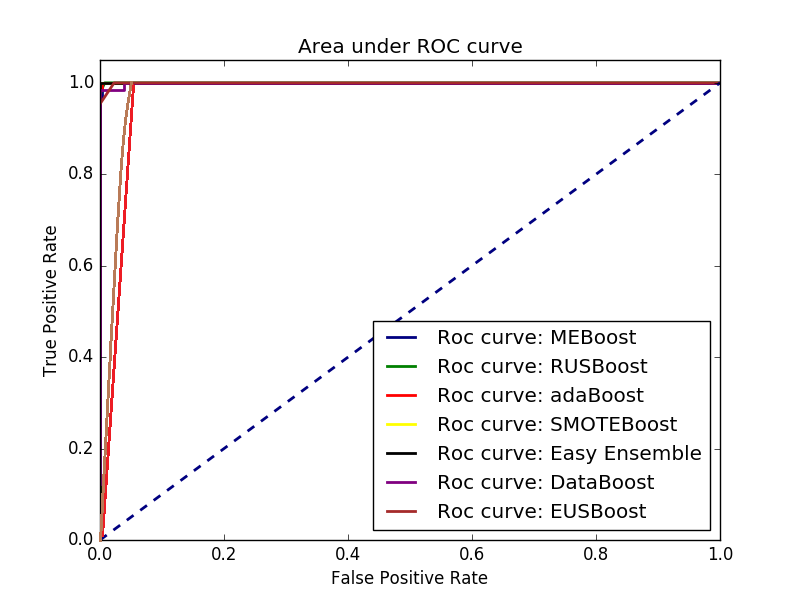}\\
(d) & (e) & (f)\\

\includegraphics [width=0.33\textwidth]{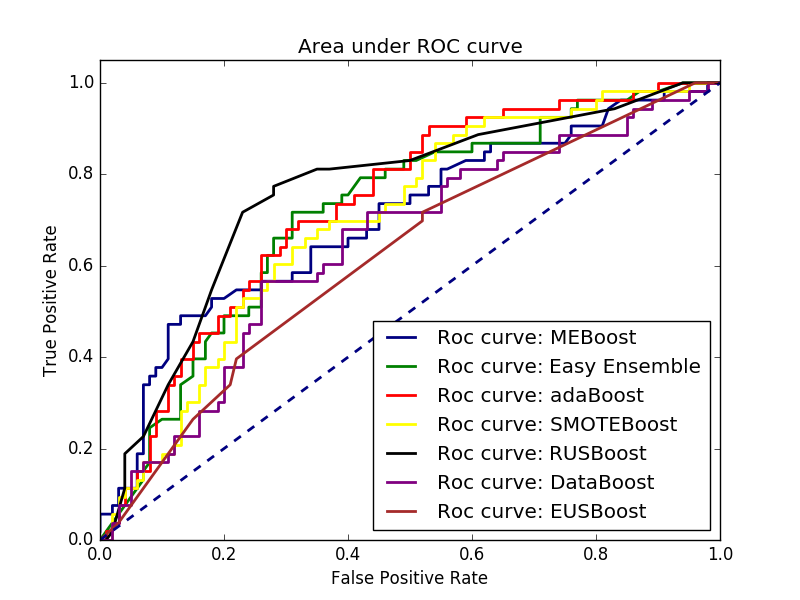}& \includegraphics [width=0.33\textwidth]{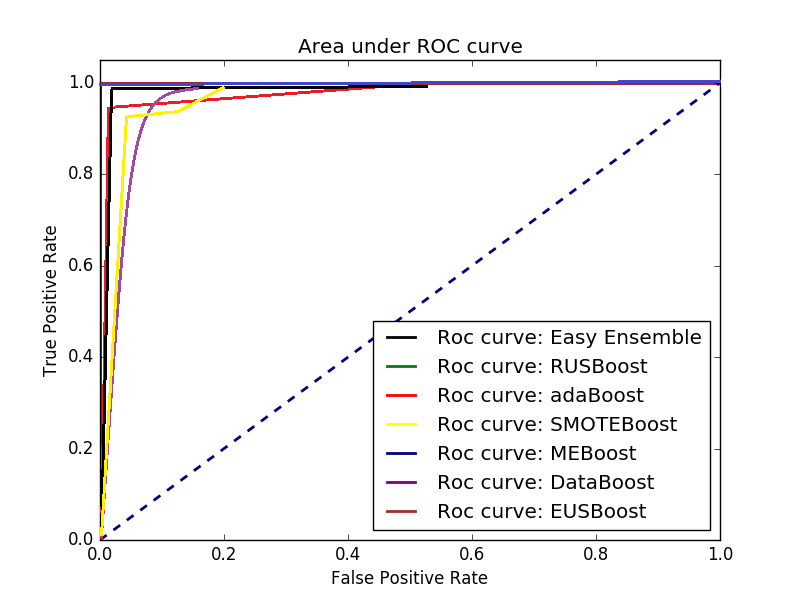}&	\includegraphics [width=0.33\textwidth]{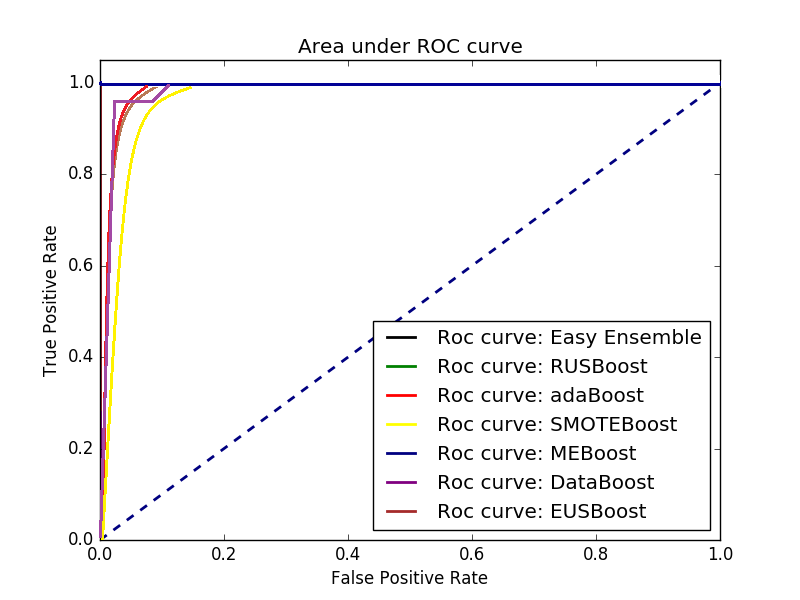}\\
(g) & (h) & (i)\\

\includegraphics [width=0.33\textwidth]{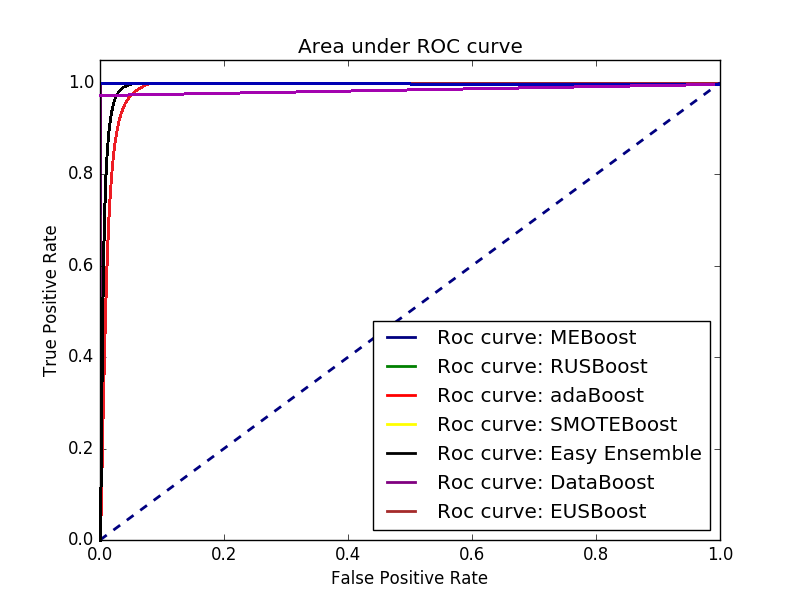}& \includegraphics [width=0.33\textwidth]{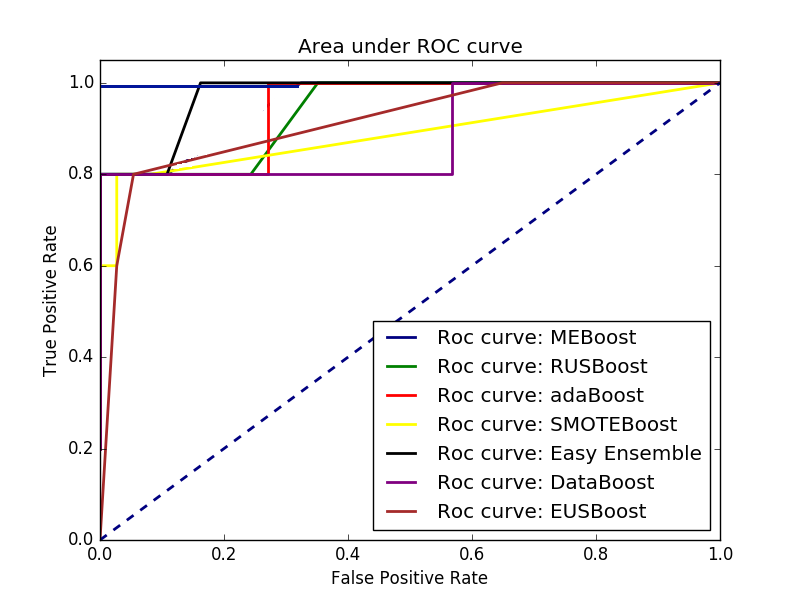}&	\includegraphics [width=0.33\textwidth]{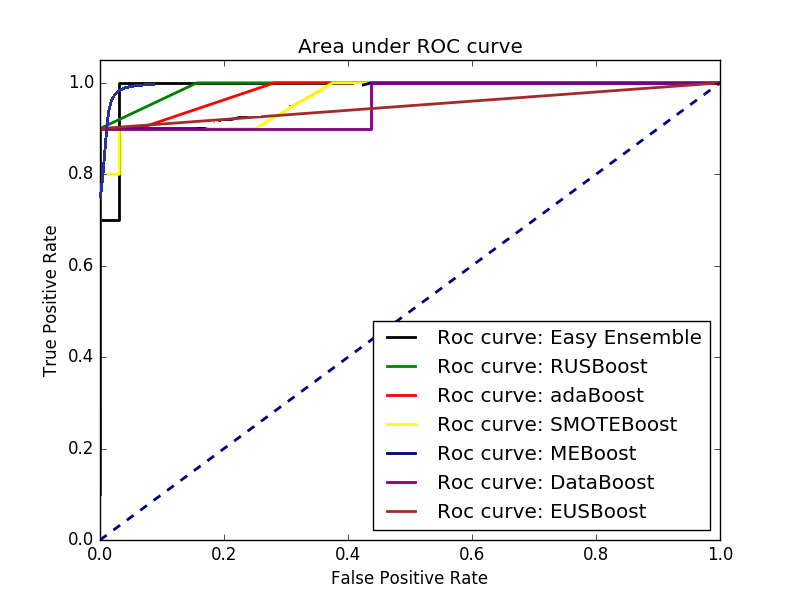}\\
(j) & (k) & (l)\\

\end{tabular}
\caption{ROC analysis for different datasets used in this paper: (a) glass5, (b) yeast6, (c )yeast5, (d) yeast4, (e) yeast-2\_vs\_4 ,(f)segment0, (g) pima, (h) page-blocks-1-3\_vs\_4, (i) newthyroid2,
(j) new-thyroid1,(k) glass6 and (l) glass-0-1-2-3\_vs\_4-5-6. \label{figRoc}}
\end{figure*}

From the tabular results from TABLE \ref{Mean_Result1}, \ref{estimator_Result} we have found that the multiple estimator technique shows better performance comparing with other state of the art boosting algorithms and other weak or strong learners as well. We also present the ROC analysis for all different datasets and the plots are given in Fig~\ref{figRoc}.

\section{Conclusion}
\label{conclusion}
Most classification algorithms primarily focuses on majority class instances rather than the minority class instances which is more important. So, the task is quite challenging to construct a classifier which can  classify minority class instances correctly. In the hope to alleviate this class imbalance problem in this paper a new algorithm called MEBoost, or Boosting with multiple learner, is presented. MEBoost has been compared with effective boosting techniques like SMOTEBoost, RUSBoost, Adaboost, DataBoost, EUSBoost and Easy Ensemble algorithms.From the experimental results, we have concluded that MEBoost performed favorably comparing with similar techniques. 

MEBoost algorithm is different from all the other boosting methods because it uses C4.5 and Extra tree classifier alternately instead of using only one of them. This allows to take advantage of both learner's characteristic and discard their individual weaknesses. Both C4.5 and Extra tree classifier have the pros and cons over one another. The results show that in using 2 different  estimators instead on 1 has much impact on auROC score.  In future, we intend to perform extensive experiments to continue investigating the performance of MEBoost with other learners.

\bibliographystyle{FILES/IEEEtran}
\bibliography{FILES/Farshid_ISIIS}

\end{document}